\documentclass[11pt]{article}

\usepackage[final]{acl}

\usepackage{times}
\usepackage{latexsym}
\usepackage{booktabs}
\usepackage[T1]{fontenc}

\usepackage[utf8]{inputenc}

\usepackage{microtype}
\usepackage{inconsolata}
\usepackage{multirow} 

\usepackage{graphicx}
\usepackage{fontawesome}

\usepackage{tikz}
\usepackage{amsmath,amssymb}
\usetikzlibrary{arrows.meta,positioning,fit,backgrounds,calc}
\usepackage{xspace}
\usepackage{cleveref}
\definecolor{cEnc}{HTML}{2C5F8A}   
\definecolor{cRec}{HTML}{2E8B74}   
\definecolor{cLin}{HTML}{C77A2B}   
\definecolor{cCRF}{HTML}{6A4C93}   
\definecolor{cLoss}{HTML}{B23A48}  
\definecolor{cOut}{HTML}{3F4A5A}   
\definecolor{cTok}{HTML}{8A93A5}   

\newcommand{\resource}[1]{\texttt{#1}}

\newcommand{\STAR}{\resource{STAR-Ar}\xspace}

\title{TTLab at Daleel 2026: \STAR, Sequence Tagging for Argument Recognition in Arabic}

\author{Bhuvanesh Verma, Ali Abusaleh, Alexander Mehler \\ 
Text Technology Lab (TTLab),  \\
         Goethe University Frankfurt\\
         \{verma,a.abusaleh,mehler\}@em.uni-frankfurt.de
         }

\begin{document}
\maketitle

\begin{abstract}
Argument Mining (AM) is a critical NLP task that remains significantly under-resourced in Arabic. 
%
This paper presents \STAR, a BERT-BiLSTM-CRF architecture for argument discourse detection and classification, as our system for Daleel 2026, the inaugural Arabic argument mining shared task. 
The task requires the identification and classification of argumentative discourse units (ADUs) in debate and editorial texts.
%
We jointly model these two objectives as a token-level sequence labeling task using a BERT-BiLSTM-CRF architecture that combines contextual transformer embeddings with structural transition constraints to support accurate span detection.
\STAR achieves an F1-score of 72.69 on validation and 73.7 on test data. 
%
Our domain-specific analysis shows that models trained exclusively on editorials underperform those trained on debates, a disparity we primarily attribute to the smaller size of the editorial dataset.
The code for \STAR is available at {\href{https://github.com/ENTAILab/daleel_2026_Arabic-Argumentative-Discourse-Mining}{\faGithub~
TTLab at Daleel 2026}}
\end{abstract}

\section{Introduction}
\label{sec:introduction}

Argument Mining (AM) is the automated extraction of argumentative structures and reasoning from natural language, with extensive applications across legal analysis, scientific literature, and public debates \cite{lawrence2019argument}. 
While various theoretical frameworks govern AM research \cite{teufel1999argumentative, walton2008argumentation}, the Toulmin Model of Argument \cite{toulmin1958uses} remains a foundational paradigm. 
In this model, the primary components are claims, premises, and warrants, with the extraction of claims and premises serving as the primary objective for most modern AM systems.

The fundamental building block in these systems is the Argumentative Discourse Unit (ADU), a term formalized by \citet{peldszus2013argument} to describe the minimal atomic text span of an argument. 
An ADU can vary significantly in length, ranging from a short dependent clause to several consecutive sentences. 
Consequently, argument extraction inherently requires both precise boundary detection, identifying the exact span of the ADU within a broader text, and subsequent classification to assign that span to a specific argumentative function (e.g., distinguishing between an author's own claim and a background claim in scientific literature).

Despite significant algorithmic advancements, the vast majority of AM research has been strictly restricted to the English language. 
Addressing this critical linguistic gap, the Daleel 2026 shared task \cite{nabhani-etal-2026-daleel} introduces a pioneering benchmark for Argument Discourse Mining in Arabic. 
The shared task is divided into two distinct but related challenges operating over the same text: 
Task 1 requires paragraph-level multi-label classification to identify all ADU types present within a given text, while Task 2 demands precise token-level span detection and classification of those ADUs.
%
We focus only on Task 2 and utilize a sequence tagging architecture.
We employ a span-based classification approach inspired by \citet{binder-etal-2022-full} to directly extract and classify ADU boundaries using contextualized representations from a pre-trained Arabic language model.
Instead of building a complex system for a severely under-resourced and imbalanced dataset, we estbalished a highly stable, mathematically constrained baseline.
%
Our proposed system, \STAR, demonstrates robust cross-domain performance on both debate and editorial texts, ultimately achieving 3rd rank among 14 participating teams in the final Daleel 2026 shared task evaluation.

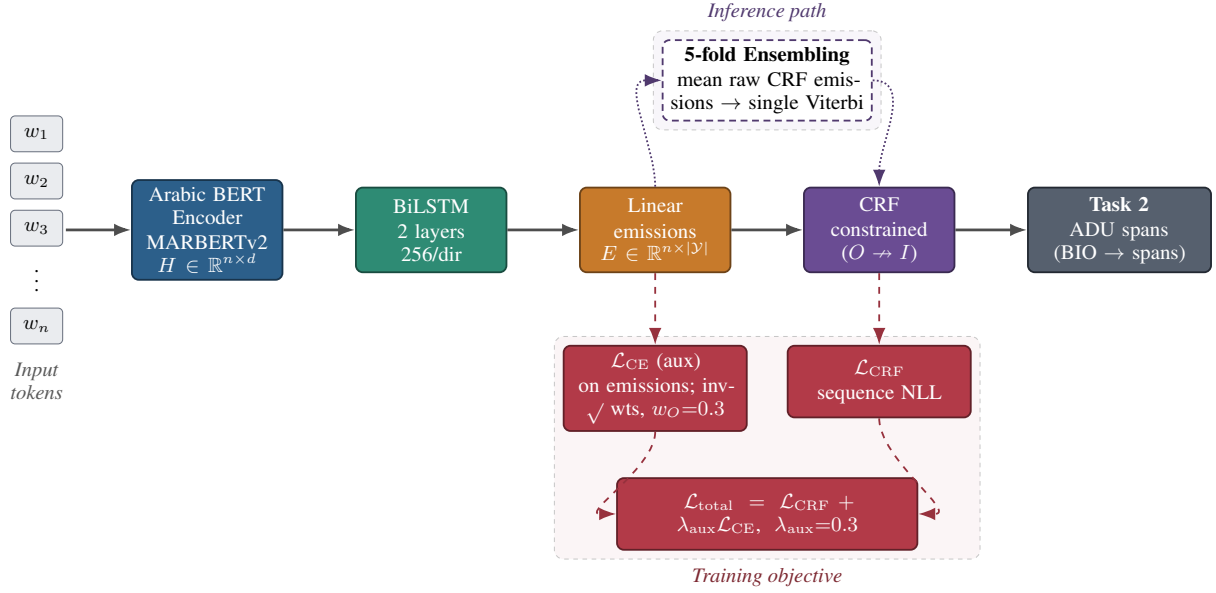
\begin{figure*}[t]
\centering
\resizebox{\textwidth}{!}{%
\begin{tikzpicture}[
  font=\small,
  >={Latex[length=2.6mm]},
  node distance=7mm and 11mm,
  block/.style={rounded corners=3pt, draw, thick, align=center, text=white,
                minimum height=13mm, text width=21mm, inner sep=3pt},
  enc/.style ={block, fill=cEnc, draw=cEnc!55!black},
  rec/.style ={block, fill=cRec, draw=cRec!55!black},
  lin/.style ={block, fill=cLin, draw=cLin!55!black},
  crf/.style ={block, fill=cCRF, draw=cCRF!55!black},
  res/.style ={block, fill=cOut, draw=cOut!55!black},
  loss/.style={block, fill=cLoss,draw=cLoss!55!black, minimum height=11mm, text width=26mm},
  ens/.style ={rounded corners=3pt, draw=cCRF!75!black, thick, fill=white, text=black,
               dash pattern=on 3pt off 2pt, align=center, minimum height=11mm, text width=30mm, inner sep=3pt},
  tok/.style ={rounded corners=2pt, draw=cTok!75!black, fill=cTok!18, text=black,
               minimum width=8mm, minimum height=5.5mm, font=\footnotesize},
  lbl/.style ={font=\footnotesize\itshape, text=black!65},
  flow/.style={->, very thick, draw=black!70},
  train/.style={->, thick, dashed, draw=cLoss!85!black},
  infer/.style={->, thick, densely dotted, draw=cCRF!75!black},
  panel/.style={rounded corners=4pt, draw=black!22, dash pattern=on 2pt off 2pt},
]

\node[tok] (t1) {$w_1$};
\node[tok, below=1.6mm of t1] (t2) {$w_2$};
\node[tok, below=1.6mm of t2] (t3) {$w_3$};
\node[below=0.2mm of t3] (td) {$\vdots$};
\node[tok, below=1.2mm of td] (tn) {$w_n$};
\node[fit=(t1)(tn), inner sep=1pt] (tokbox) {};
\node[lbl, below=1mm of tokbox.south, align=center] {Input\\tokens};

\node[enc, right=10mm of tokbox] (bert)
  {Arabic BERT Encoder\\[1pt]\footnotesize MARBERTv2\\ $H\in\mathbb{R}^{n\times d}$};
\node[rec, right=of bert] (lstm)
  {BiLSTM\\[1pt]\footnotesize 2 layers\\ 256/dir};
\node[lin, right=of lstm] (lin)
  {Linear\\[1pt]\footnotesize emissions\\ $E\in\mathbb{R}^{n\times|\mathcal{Y}|}$};
\node[crf, right=of lin] (crf)
  {CRF\\[1pt]\footnotesize constrained\\ ($O\!\nrightarrow\!I$)};
\node[res, right=of crf, text width=26mm, fill=cOut!88] (span)
  {\textbf{Task 2}\\[1pt]\footnotesize ADU spans\\ (BIO $\to$ spans)};

\draw[flow] (tokbox.east) -- (bert.west);
\draw[flow] (bert) -- (lstm);
\draw[flow] (lstm) -- (lin);
\draw[flow] (lin)  -- (crf);
\draw[flow] (crf)  -- (span);

\coordinate (imid) at ($(lin.north)!0.5!(crf.north)$);
\node[ens, above=10mm of imid] (ensb)
  {\textbf{5-fold Ensembling}\\[1pt]\footnotesize mean raw CRF emissions $\to$ single Viterbi};
\draw[infer] (lin.north) to[out=90,in=180] (ensb.west);
\draw[infer] (ensb.east) to[out=0,in=90] (crf.north);

\node[loss, below=11mm of lin] (lce)
  {$\mathcal{L}_{\mathrm{CE}}$ (aux)\\[1pt]\footnotesize on emissions; inv-$\sqrt{}$ wts, $w_O{=}0.3$};
\node[loss, below=11mm of crf] (lcrf)
  {$\mathcal{L}_{\mathrm{CRF}}$\\[1pt]\footnotesize sequence NLL};
\coordinate (lmid) at ($(lce.south)!0.5!(lcrf.south)$);
\node[loss, below=8mm of lmid, text width=44mm] (ltot)
  {$\mathcal{L}_{\mathrm{total}}=\mathcal{L}_{\mathrm{CRF}}+\lambda_{\mathrm{aux}}\mathcal{L}_{\mathrm{CE}}$,\ \ $\lambda_{\mathrm{aux}}{=}0.3$};
\draw[train] (lin.south) -- (lce.north);
\draw[train] (crf.south) -- (lcrf.north);
\draw[train] (lce.south) to[out=-90,in=180] (ltot.west);
\draw[train] (lcrf.south) to[out=-90,in=0] (ltot.east);

\begin{scope}[on background layer]
  \node[panel, fill=cCRF!5, fit=(ensb),
        label={[lbl,text=cCRF!75!black]above:Inference path}] {};
  \node[panel, fill=cLoss!5, fit=(lce)(lcrf)(ltot),
        label={[lbl,text=cLoss!70!black]below:Training objective}] {};
\end{scope}

\end{tikzpicture}%
}
\caption{Overview of \STAR, a BERT--BiLSTM--CRF sequence tagger for Task 2 (ADU span detection and classification) of Daleel 2026.}
\label{fig:arch}
\end{figure*}

The paper is organized as follows: \Cref{sec:backgorund} provides the task description, dataset detail, and related work.
\Cref{sec:sys_overview} provides the details regarding the proposed system architecture.
%
In \Cref{sec:exp_setup}, we present the experimental setup for the shared task.
\Cref{sec:res_and_dis} provides a comprehensive analysis of our experimental results alongside ablation studies. 
Finally, \Cref{sec:conclusion} concludes the paper.


\section{Background}
\label{sec:backgorund}

Argument mining (AM) focuses on automatically identifying and extracting reasoning structures from natural language, typically through sentence-level classification or span-level sequence tagging paradigms.
Although this task have been widely studied in English, Arabic remains under-resourced.
To this end, the Daleel 2026 introduced two shared tasks which focuses on the identification and classification of argumentative discourse units (ADUs) in Arabic Text.

\begin{table*}[t]
\centering
\small
\begin{tabular}{llccccccc}
\toprule
\textbf{Split} & \textbf{Genre} & \textbf{Paragraphs} & \textbf{AS} & \textbf{OT} & \textbf{AN} & \textbf{TE} & \textbf{CO} & \textbf{ST} \\
\midrule
\multirow{3}{*}{Train}
& Overall    & 612 & 462 (75.5\%) & 222 (36.3\%) & 182 (29.7\%) & 160 (26.1\%) & 36 (5.9\%) & 28 (4.6\%) \\
& Debate     & 357 & 287 (80.4\%) & 202 (56.6\%) &  69 (19.3\%) & 100 (28.0\%) & 22 (6.2\%) &  6 (1.7\%) \\
& Editorial  & 255 & 175 (68.6\%) &  20 (7.8\%)  & 113 (44.3\%) &  60 (23.5\%) & 14 (5.5\%) & 22 (8.6\%) \\
\midrule
\multirow{3}{*}{Dev}
& Overall    & 217 & 157 (72.4\%) &  63 (29.0\%) &  68 (31.3\%) &  76 (35.0\%) & 12 (5.5\%) &  9 (4.1\%) \\
& Debate     & 129 & 106 (82.2\%) &  52 (40.3\%) &  33 (25.6\%) &  43 (33.3\%) &  7 (5.4\%) &  1 (0.8\%) \\
& Editorial  &  88 &  51 (58.0\%) &  11 (12.5\%) &  35 (39.8\%) &  33 (37.5\%) &  5 (5.7\%) &  8 (9.1\%) \\
\bottomrule
\end{tabular}
\caption{Paragraph-level label coverage in the Daleel 2026 dataset. Each value reports the number and percentage of paragraphs containing at least one span of the corresponding label. Percentages are computed with respect to the number of paragraphs in each row and do not sum to 100\% because a paragraph may contain multiple labels.}
\label{tab:task2_para_stats}
\end{table*}

\subsection{Task 1: ADU Paragraph Classification}
Task 1 is a multi-label classification challenge. 
%
Given a paragraph, the system must predict the presence of one or more ADU classes: \textit{Common ground}, \textit{Assumption}, \textit{Testimony}, \textit{Statistics}, \textit{Anecdote} and \textit{Other}. 
The training labels exhibit strong class imbalance; for instance, \textit{Assumption} is the most frequent label (appearing in 75.5\% of training paragraphs), while \textit{Statistics} is the rarest (4.6\%). 
On average, each paragraph is annotated with 2.0 labels.

\subsection{Task 2: ADU Span Detection}
Task 2 is a token-level sequence tagging task. Given a paragraph, systems must identify the exact character span of an ADU and classify it into its corresponding class. 
A single text can contain multiple ADU spans of the same or different classes. 
The training set consists of 2,975 ground-truth spans with exceptionally dense coverage, 84.6\% of training paragraphs have between 80\% and 100\% of their text volume bound to a labeled span.

The dataset for the shared tasks is partitioned into a training set ($n=612$), a development set ($n=217$) and a final evaluation test set ($n=213$). 
All splits share highly consistent domain distributions. 
However, distribution of labels although fairly consistent but highly skewed in both train and dev set.
Common grounds and Statistics have the least amount of data instances (see \autoref{tab:task2_para_stats}).

We worked exclusively with task 2 under closed track rules which is essentially a sequence tagging task.
Early sequence-based methods in argument mining used conditional random fields (CRFs) over BIO-encoded tokens to identify premises \citep{goudas2014argument}.
This approach was later extended to the joint extraction of major claims, claims, and premises from text \citep{stab2017parsing}.
To capture complex long-range sequential dependencies more effectively, modern systems use deep learning and contextualized language models for end-to-end span extraction.
\citet{eger2017neural} advanced neural end-to-end argument mining by investigating BiLSTM-CRF sequence tagging models for identifying exact argument boundaries.
Building on this line of work, \citet{binder-etal-2022-full} introduced a BERT-BiLSTM-CRF architecture for identifying and classifying scientific argumentative discourse units (ADUs).
Because the Daleel 2026 shared task requires precise extraction and classification of ADUs within Arabic texts, the framework established by \citet{binder-etal-2022-full} serves as the foundational basis for our methodology. 
\section{System Overview}
\label{sec:sys_overview}
To account for the highly imbalanced and structured nature of ADU detection and classification, we formulate the problem as a sequence tagging task using a BIO (begin, inside, outside) tagging scheme.
%
For this task, we propose \STAR, a pre-trained transformer architecture augmented with a recurrent layer and a structured prediction head.

\subsection{Model Architecture}
The core architecture uses a BERT-based text encoder pre-trained on Arabic corpora to extract contextualized representations of the input sequence.
%
To capture longer-range dependencies and sequential argumentative structures, the transformer's final hidden states are passed to a bidirectional long short-term memory (BiLSTM) network.
%
A linear classifier then projects the BiLSTM outputs into the BIO-encoded label space, producing emission scores.
%
Finally, a conditional random field (CRF) layer is applied on top of these emission scores.
%
To guarantee the structural validity of the predicted boundaries, we explicitly initialize the CRF transition matrix to heavily penalize invalid BIO sequences. 
Invalid paths such as sequence initializations starting directly with an \texttt{I}-tag, \texttt{O} $\rightarrow$ \texttt{I} transitions, and cross-label continuations (e.g., \texttt{B-AS} $\rightarrow$ \texttt{I-TE}), are assigned a mathematically prohibitive transition score ($-10000.0$). 
This strictly forces the Viterbi decoder to output structurally coherent label sequences.


\subsection{Optimization and Loss Formulation}
To directly counteract the severe class imbalance and density of the dataset, STAR-Ar bypasses standard uniform optimization in favor of a custom, cost-sensitive composite loss formulation.  
The primary objective is the negative log-likelihood of the CRF sequence ($L_{CRF}$). 
We supplement this with an auxiliary token-level Cross-Entropy loss ($L_{CE}$) applied directly to the BiLSTM emissions. 
The CE loss utilizes inverse-square-root class weights, with the \textit{O} tag weight capped at 0.3 to prevent it from overwhelming the ADU classes. 
The final loss is defined as:
$$L_{total}=L_{CRF}+\lambda_{aux}L_{CE}$$
where $\lambda_{aux}$ is an auxiliary weighting parameter.

\section{Experimental Setup}
\label{sec:exp_setup}
To ensure robust evaluation and generalization, we utilize a 5-fold stratified cross-validation strategy. 
Stratification is based on the least frequent label present within a document to guarantee that rare classes are distributed evenly across folds. 
During training, we apply a weighted random sampler at the document level to over-sample paragraphs containing minority ADU classes.

To determine the optimal contextual representation for this task, we evaluated three distinct pre-trained Arabic text encoders. 
We found \resource{MARBERTv2} consistently outperformed both \resource{AraBERT} variants across multi seed experiments, so we used \resource{MARBERTv2} as text encoder for the final system (see \autoref{app:encoder_selection}).
Regarding other hyperparameters, we apply a transformer dropout rate of 0.1, and the two-layer BiLSTM utilizes a hidden size of 256 per direction. 
The network is optimized using AdamW for up to 20 epochs with a patience of 5 epochs for early stopping. 
We employ differential learning rates: $2 \times 10^{-5}$ for the BERT encoder and $1 \times 10^{-3}$ for the randomly initialized top layers (BiLSTM, CRF, and linear heads), alongside a linear learning rate scheduler with a 10\% warm-up phase and gradient clipping at 1.0.

The final system for test evaluation is trained with both train and dev set using same 5-fold stratifed cross validation strategy.
%
For predictions, we leverage an ensembling technique at the emission level. 
%
Rather than applying majority voting to the final decoded label sequences, we compute token-wise averages of the raw CRF emission probabilities across all trained folds and perform a single Viterbi decoding step on the averaged emissions to produce the final predicted spans.

\begin{table}[t]
\centering
\small
\begin{tabular}{llccc}
\toprule
\textbf{Split} & \textbf{Evaluation} & \textbf{Editorial} & \textbf{Debate} & \textbf{Both} \\
\midrule
\multirow{3}{*}{Dev}
& Editorial & 60.80 & 51.30 & \textbf{66.52} \\
& Debate    & 58.28 & \textbf{75.29} & 75.04 \\
& Both      & 60.15 & 69.25 & \textbf{72.69} \\
\midrule
\multirow{3}{*}{Test}
& Editorial & 62.35 & 54.52 & \textbf{63.56} \\
& Debate    & 60.07 & 76.17 & \textbf{77.84} \\
& Both      & 61.25 & 70.68 & \textbf{73.74} \\
\bottomrule
\end{tabular}
\caption{F1-score (\%) for evaluation on the Daleel development and test sets. Columns denote the training domain, while rows denote the evaluation split and domain.}
\label{tab:cross_domain_macrof1}
\end{table}

\section{Results and Discussion}
\label{sec:res_and_dis}

\autoref{tab:cross_domain_macrof1} presents \STAR's performance on the development set as well as test set for Task 2.
%
Overall, we see model performing better on test set which is likely because we added development data to train the final system.
Furthermore, we observe that the model trained on both domains performs better on debates than on editorials which is understandable given the distribution of data across domains.
Training on debate domain alone yields performance close to training on the both domains, suggesting that debate data captures many of the linguistic patterns needed for ADU identification.
%
%
On the other hand, models trained only on editorials benefit more from incorporating debate data, likely because the editorial subset is smaller and exhibits greater label imbalance. 
Consequently, the additional debate examples provide useful supervision that improves generalization to editorials.

\begin{figure}[t]
    \centering
    \includegraphics[width=\linewidth]{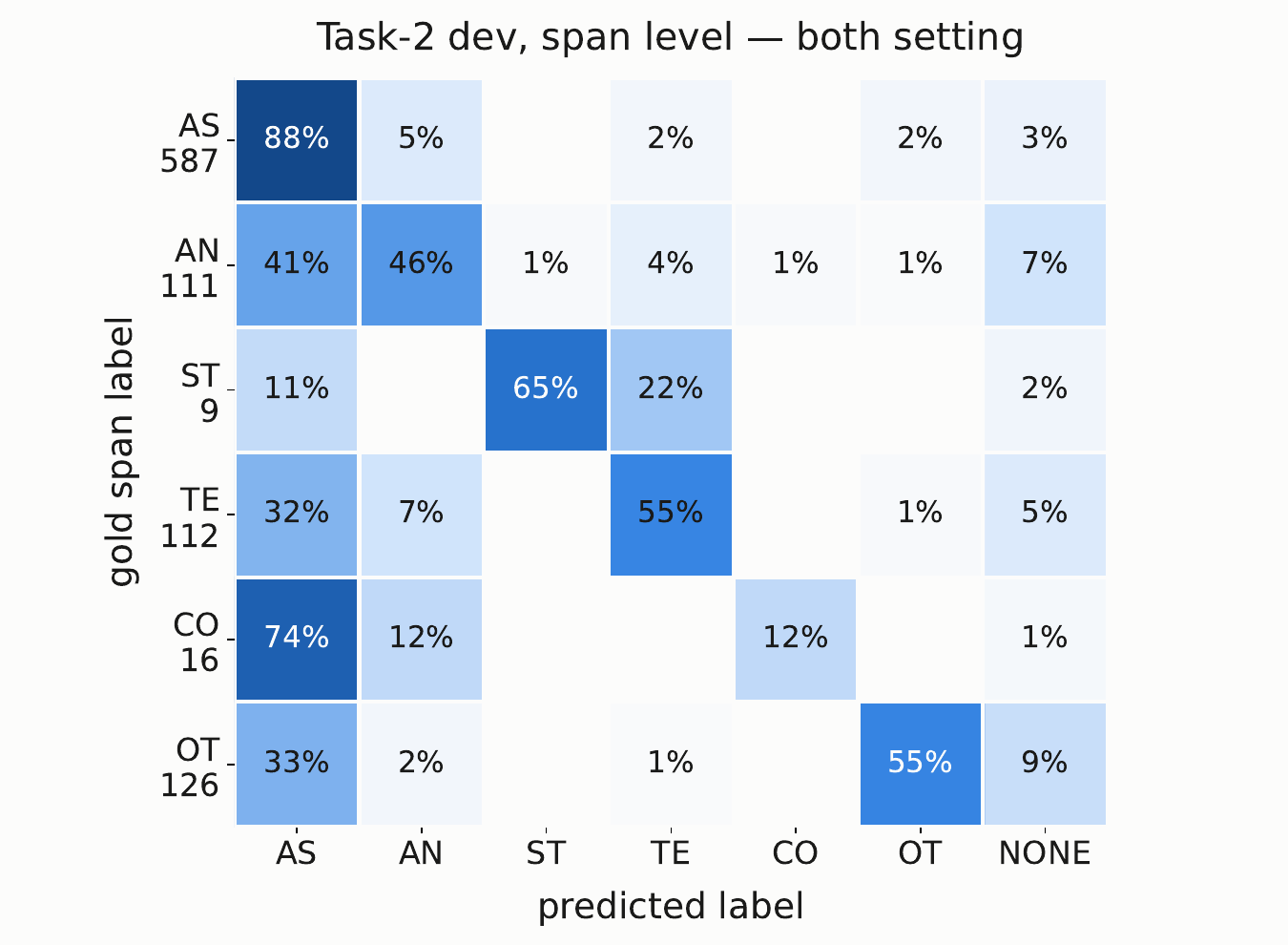}
     \caption{Cross-domain span-level confusion matrix. Rows are gold classes; cells show mean character-wise overlap with predictions. The diagonal represents span-level recall; 'NONE' captures false negatives.}
    \label{fig:confusion_matrix}
\end{figure}

As detailed in \autoref{tab:task2_para_stats}, the dataset exhibits severe class imbalance. 
To evaluate the system's span-level boundary resolution across these skewed distributions, we analyze the character-overlap confusion matrix in \autoref{fig:confusion_matrix}.
Overall, the model exhibits a strong predictive bias toward the majority class, Assumption (AS). 
While AS achieves the highest recall (88.49\%), the model systematically overpredicts this label across all other categories. 
This absorption effect is most severe for Common ground (CO), where 74.22\% of gold characters are erroneously captured by AS predictions, reducing CO's recall to just 12.26\%.
Mid-frequency classes such as Anecdote (AN), Testimony (TE), and Other (OT) demonstrate moderate robustness, maintaining recall rates between 45\% and 55\%. 
However, they still lose a substantial portion of their span volume (ranging from 31\% to 40\%) to the majority AS class.
Interestingly, despite being the rarest class, Statistics (ST) performs robustly as the second-best category with a recall of 64.77\%. 
Because its explicit quantitative surface features provide a highly separable linguistic signal, it largely resists absorption into the majority class, with its primary misclassifications splitting between TE (22.06\%) and AS (11.11\%).

\section{Conclusion}
\label{sec:conclusion}
In this work, we present \textbf{STAR-Ar}, a BERT-BiLSTM-CRF framework tailored for argumentative discourse unit (ADU) extraction and classification in Arabic. 
On the shared task final test set, our system achieved a F$_1$ score of 73.74 on the both domain setting, ranking 3rd out of 14 participating teams. 
Our findings demonstrate that sequence labeling remains a competitive, highly efficient paradigm for joint span detection and classification in argument mining.
Qualitative analysis reveals that while class re-balancing yields non-uniform performance gains, label frequency alone fails to predict class-wise difficulty.
Finally, cross-domain evaluations indicate a high sensitivity to domain shifts, highlighting the necessity for larger, more structurally diverse annotated corpora to further advance Arabic argument mining.

\section*{Limitations}
We used small language models with 512-subword limit which truncates the long inputs. 
Although we verified that it affect negligible portion of the dataset (1.14\% in train, 0.46\% in dev and 0.47\% in test), but it still constrains applicability of our system to longer documents.
Another limitation of this work is its generalization capabilities across other Arabic text types and dialectical variations.

\section*{Acknowledgments}
This research is funded by the German Research Foundation within the project \emph{(Semi-)Automated thematic text classification as a basis for corpus-linguistic value-added services} (Project number: \href{https://gepris.dfg.de/project/531750631}{531750631}) and within the Infrastructure Priority Programme \emph{New Data Spaces for the Social Sciences} \href{https://www.dfg.de/de/aktuelles/neuigkeiten-themen/info-wissenschaft/2023/info-wissenschaft-23-20}{(SPP 2431)}, Research-driven Infrastructure for Advanced Survey-related Data (CIRCLET) measure project number ~\href{https://gepris.dfg.de/project/539634240}{539634240}.

\bibliography{custom}

\appendix

\section{Encoder Comparison}
\label{app:encoder_selection}

In order to decide which encoder is best for our system, we performed multi seed experiment using three different Arabic encoders. We ran same 5 fold stratified cross validation on train set across three different seeds to obtain results shown in \autoref{tab:ablation_encoders_pct}.
\begin{table}[h]
\centering
\resizebox{\columnwidth}{!}{%
\begin{tabular}{lc}
\toprule
\textbf{Encoder Model} & \textbf{Task 2 Overall $F_1$ (\%)} \\
\midrule
AraBERTv02-Twitter-b & $63.75 \pm 0.43$ \\
AraBERTv02-Twitter-l & $65.82 \pm 2.17$ \\
MARBERTv2            & $67.53 \pm 1.53$ \\
\bottomrule
\end{tabular}}
\caption{Ablation Study: Impact of different pre-trained text encoders on overall performance (models trained on the combined dataset).}
\label{tab:ablation_encoders_pct}
\end{table}




\section{Paragraph Level Error Analysis}

\begin{figure}[t]
    \centering
    \includegraphics[width=\linewidth]{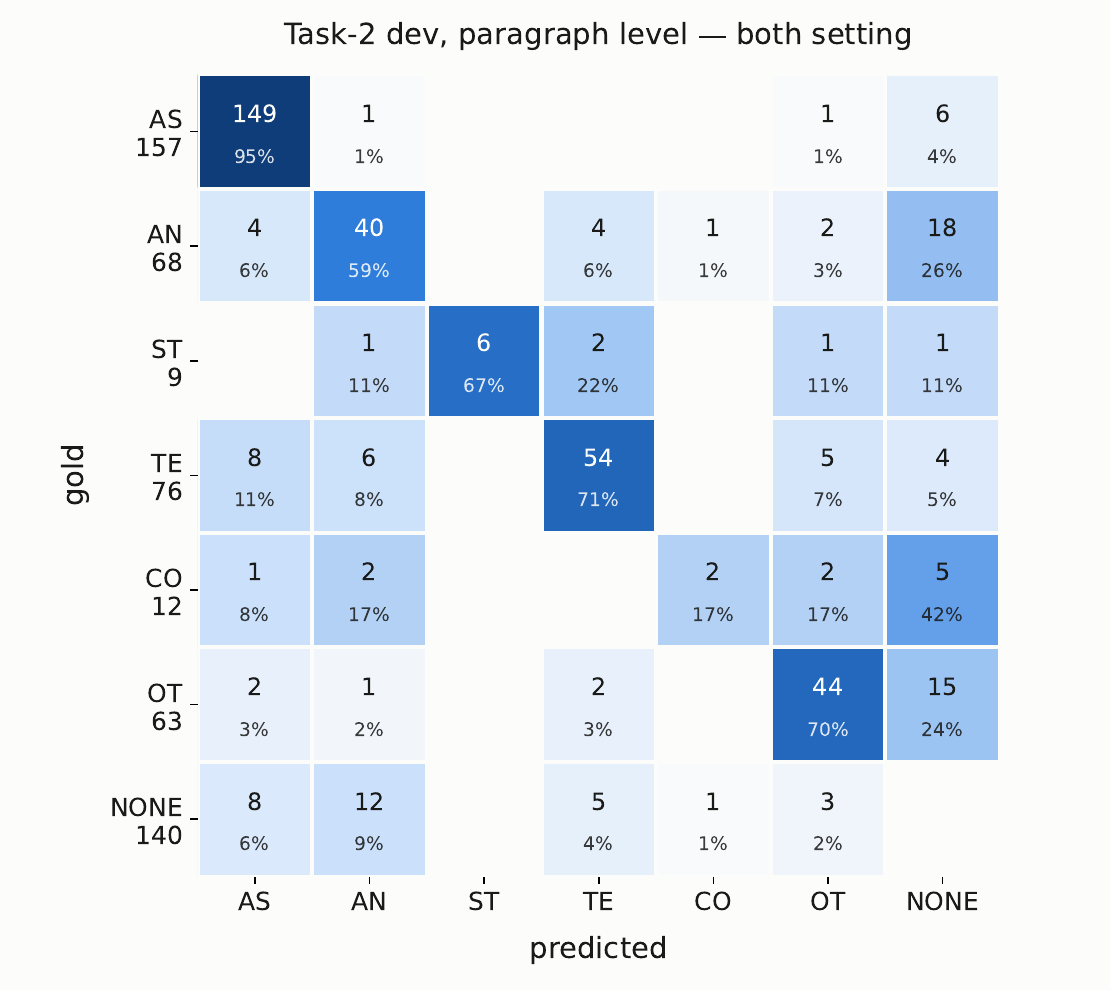}
     \caption{Paragraph-level dev-set heatmap (cross-domain trained). Rows list gold labels with paragraph support; the diagonal gives recall. Off-diagonal cells show spurious predictions on missed gold paragraphs.}
    \label{fig:confusion_matrix_para}
\end{figure}

To evaluate the efficacy of our proposed system for the class imbalance, we analyze the development set confusion matrix (\autoref{fig:confusion_matrix_para}). 
Overall, the model robustly identifies the majority class, Assumption (AS), alongside mid-frequency classes, Testimony (TE) and Other (OT). 
Misclassifications among minority classes are predominantly driven by false negatives (omissions) rather than cross-label confusion.
Specifically, Anecdote (AN) and Common ground (CO) are most frequently misclassified as None rather than assigned competing labels.

Despite being the rarest class, Statistics (ST) achieves 100\% precision (recovering 6 of 9 instances without false positives), as its explicit quantitative surface features provide a highly separable signal that cost-sensitive weighting exploits. 
Conversely, CO yields the lowest recall (2 of 12 instances), indicating that class-re-balancing techniques fail to surface minority classes that lack distinct lexical or structural cues. 
This pattern remains consistent in single-domain configurations, underscoring the necessity of multi-domain training which mitigates cross-domain sparsity by leveraging complementary label distributions.

\end{document}